\documentclass[runningheads]{llncs}
\usepackage[T1]{fontenc}
\usepackage{graphicx}
\usepackage{booktabs}
\usepackage[misc]{ifsym}
\usepackage{amsmath}
\usepackage{amssymb}
\usepackage{bbm}
\usepackage{booktabs,longtable,xcolor}
\usepackage{subcaption}
\usepackage{float}
\newcommand{\corr}{(\Letter)}
\usepackage{mwe}

 \usepackage{multirow}

\newcommand{\wga}{WG-TNR}

\begin{document}

\title{Face Age Verification  Vulnerabilities Under 
Simple Appearance Manipulations}

\titlerunning{Accepted at the BIAS Workshop, ECML PKDD 2026}

\author{Ioannis Sarridis \corr \and
Ioannis Kompatsiaris \and
Symeon Papadopoulos}
\authorrunning{I. Sarridis et al.}
\institute{Information Technologies Institute, Centre for Research and Technology Hellas, 6th km Charilaou-Thermi Rd, Thessaloniki, 57001, Greece
\email{\{gsarridis,ikom,papadop\}@iti.gr}}




\maketitle              

\begin{abstract}
Online platforms increasingly rely on automated age estimation systems to enforce minimum-age policies. Focusing on vision-based models designed for this task, concerns arise regarding their robustness to simple appearance changes that underage individuals may use to bypass such systems, such as drawing a mustache or applying lipstick. In this work, we present a systematic study of age verification robustness by simulating visual alterations that can be easily achieved by underage individuals. We evaluate seven models, including vision, vision-language, and multimodal large language models, across three datasets and four manipulation types. Interestingly, under drawn beard stubble, up to 61\% of True Negatives are flipped into False Positives. Furthermore, we investigate how different demographics are affected by such manipulations, finding that Indians are more affected by beard stubble manipulations, while females are more affected than males across all manipulations. Finally, we explore how these biases can be mitigated using bias mitigation methodologies in lightweight linear probe settings.

\end{abstract}
\begin{figure}[t]
    \centering
    \includegraphics[width=\linewidth]{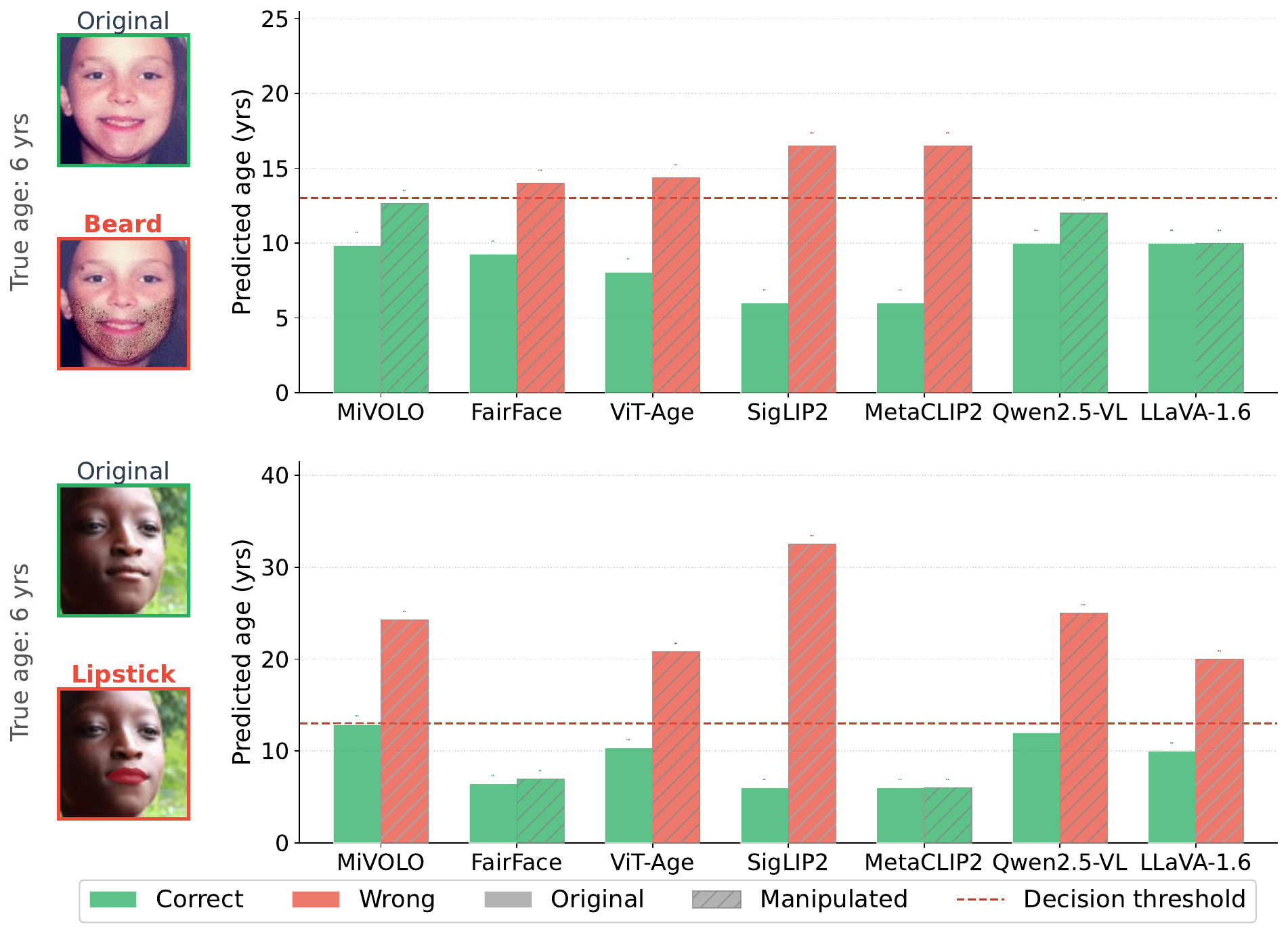}
    \caption{A single straightforward appearance manipulation can be sufficient to make age verification models classify a 6-year-old as at least 13 years old. Adding drawn beard stubble (top) causes 4 of 7 models to cross the verification threshold, while applying red lipstick (bottom) causes 5 of 7 models to do so.}
    \label{fig:teaser}
\end{figure}

\section{Introduction}

Computer vision models are widely used in decision making pipelines where the output depends on personal and often sensitive visual attributes, such as identity, gender, and age~\cite{buolamwini2018gender,karkkainen2021fairface}. In these settings, models may perform well on data that share the same distribution as the training data, but fail when a user applies superficial appearance changes~\cite{geirhos2020shortcut}. This matters most when the model is part of an access-control system, because the user has a direct incentive to induce a favorable prediction.

Spurious correlations present within the training data often drive this behavior of deep learning models~\cite{geirhos2020shortcut,sagawa2020groupdro,sarridis2025mavias,sarridis2025vb}.
A spurious attribute is predictive of the target label in the training data, but it is not causally tied to the target concept. In face analysis, such attributes include hair color, makeup, cosmetics, etc. These attributes can be easier to learn than the actual target attribute, and when they co-occur strongly with the target labels, models tend to rely on them to make decisions~\cite{nam2020lff,sagawa2020groupdro}.
For instance, training a gender classifier using the CelebA \cite{celeba} dataset, where blond hair is strongly correlated with females, results in a model that exploits hair color as a shortcut to predict gender \cite{sagawa2020groupdro,sarridis2025badd}.
On the other hand, some attributes have a causal relation to the target, but relying on them could create safety issues.
For age estimation, facial hair is such an example, since it is highly correlated with adulthood, so a model can treat it as strong evidence of age. However, given that this attribute is easy to modify, an underage individual can draw beard stubble or a mustache, so that a model that has learned to rely on this attribute increases its predicted age.

Online platforms use automated age assurance to enforce minimum-age policies, including the common 13-year threshold used by services such as Instagram and Facebook \cite{meta2026age}. Such systems combine several signals, including both visual information and network interactions. Focusing on face-based age estimation, cases of underage individuals bypassing online age verification with a fake mustache have already been reported \cite{wired2026age}, raising the need for systematic evaluation under simple manipulations.

In this work, 
we present a systematic study of age verification robustness under digitally applied facial manipulations that simulate simple appearance cues.
We evaluate seven models across three datasets and four deterministic, landmark-aligned manipulations, namely beard stubble, mustache, eye makeup, and lipstick. The models considered in the comparative analysis include vision models, vision-language models (VLMs), and multimodal large language models (MLLMs). Results show that most tested models are highly vulnerable to such manipulations, e.g., drawn beard stubble flips up to 61\% of correctly classified underage predictions. 
This vulnerability is substantially less pronounced for LLaVA-1.6, which is consistently the least affected model in our evaluation.
Moreover, it is worth noting that the effect is not uniform across demographic groups, e.g., female subjects are affected more than males across all the manipulations. Finally, we show that mitigation with linear probes on frozen features can improve robustness with only small amounts of manipulated images within training data.
Our contributions are as follows:
\begin{itemize}
    \item  A study of age verification robustness under digitally applied facial manipulations that simulate simple attack scenarios by underage individuals. 
    
    \item A manipulation-induced vulnerability analysis across demographic groups, showing that the risk is unevenly distributed across race and gender.
    
    \item A comparative analysis showing how bias mitigation approaches can increase the models' robustness under lightweight linear probe settings with limited manipulated samples within the training data.
\end{itemize}

\section{Related Work}

\paragraph{Facial age estimation and age verification.}
Facial age estimation has been studied as both a regression task, where a model predicts a continuous age, and a classification task, where the face is assigned to a discrete age group~\cite{eidinger2014age,rothe2015dex}. 
Early approaches relied on hand-crafted shape, texture, or aging-pattern representations, while recent systems use deep convolutional networks, transformers, and vision-language representations trained on large in-the-wild face datasets~\cite{rothe2015dex,kuprashevich2023mivolo,li2022ordinalclip}. 
MiVOLO combines face and body information through a multi-input transformer for joint age and gender estimation, improving robustness when the face is partially degraded or occluded~\cite{kuprashevich2023mivolo}. 
OrdinalCLIP formulates age estimation as language-guided ordinal regression by learning rank-aware prompts over CLIP features~\cite{li2022ordinalclip}. 
Face-attribute models such as FairFace jointly predict age, gender, and race, and were introduced together with a demographically balanced dataset to reduce evaluation artifacts caused by skewed face datasets~\cite{karkkainen2021fairface}.

\paragraph{Physical and appearance-based attacks on face analysis.}
Sharif et al.~\cite{sharif2016accessorize} showed that printed adversarial eyeglass frames can induce dodging and impersonation errors in face recognition systems, establishing one of the first practical physical adversarial attacks on face recognition. 
Komkov et al.~\cite{komkov2021advhat} later proposed AdvHat, where an adversarial pattern printed on a hat attacks ArcFace under real-world capture conditions. 
Other patch-based methods generate adversarial regions for dodging or impersonation, often through optimization or generative models~\cite{hwang2023adversarialpatch}. 
Light- and projection-based attacks similarly manipulate the captured facial image through external physical signals rather than natural appearance changes~\cite{duan2021laser,liu2025projattacker}.

\paragraph{Shortcut learning and visual spurious correlations.}
In vision models, shortcuts arise from co-occurring visual attributes that are easier to learn than the intended concept~\cite{geirhos2020shortcut,pezeshki2021gradient}. 
This issue is especially important in face analysis because facial attributes are statistically entangled: hair, cosmetics, skin texture, facial hair, age, gender presentation, and capture conditions often co-occur in natural image datasets~\cite{karkkainen2021fairface,buolamwini2018gender}. 
Several methods have been proposed to improve robustness under group shifts and spurious correlations. 
GroupDRO optimizes the worst-case loss over predefined groups and is effective when group annotations are available~\cite{sagawa2020groupdro}. 
JTT reduces the need for full group labels by first training an ERM model and then upweighting examples misclassified by that first model~\cite{liu2021jtt}. 
LfF trains a biased model and uses its failures to guide a debiased classifier toward samples that conflict with dominant shortcuts~\cite{nam2020lff}.
FLAC minimizes mutual information between the spurious attributes and the target attributes using an auxiliary bias-capturing model~\cite{sarridis2024flac}.
BPA derives pseudo-attributes via unsupervised feature-space clustering and applies cluster-wise reweighting to expose bias-conflicting samples without requiring explicit group labels~\cite{bpa2024}.
These methods motivate our mitigation study, where we evaluate whether lightweight linear probes on frozen representations can improve robustness to manipulation-induced group shifts without retraining the full backbone.

\paragraph{Demographic bias in face analysis.}
Face analysis systems have repeatedly shown uneven performance across demographic groups. 
Gender Shades~\cite{buolamwini2018gender} demonstrated large intersectional disparities in commercial gender classification, with the highest error rates for darker-skinned women and much lower error rates for lighter-skinned men. 
Similar disparities have also been reported in face verification, where an in-depth analysis of intersectional biases across race, age, and gender showed substantially higher error rates for specific demographic subgroups, such as Asian women~\cite{sarridis2023towards}.
FairFace addressed part of this issue by collecting a more balanced face attribute dataset in terms of race, gender, and age, and showed that training data composition affects generalization across demographic groups~\cite{karkkainen2021fairface}. 
These findings established that aggregate accuracy can hide large subgroup failures in facial analysis~\cite{buolamwini2018gender,karkkainen2021fairface}.

\section{Benchmark Setup}
\label{sec:setup}

\subsection{Problem Formulation}
\label{sec:notation}

Let $f: \mathcal{X} \rightarrow \mathbb{R}$ denote an age estimator that maps a face image
$x \in \mathcal{X}$ to a predicted age $\hat{y}=f(x)$.
Age verification at threshold $\tau$ converts this continuous prediction into a binary decision:
\[
h_f(x;\tau)=\mathbbm{1}[f(x)\geq \tau],
\]
where $h_f(x;\tau)=0$ denotes an underage prediction and $h_f(x;\tau)=1$ denotes an age-eligible prediction.
We use $\tau=13$ throughout.

We study how this decision changes under very simple appearance manipulations.
Let $\mathcal{T}_m:\mathcal{X}\rightarrow\mathcal{X}$ denote a manipulation operator of type
$m \in \mathcal{M}$, such as drawn beard stubble, lipstick, eye makeup, or mustache.
For an input image $x$, the corresponding manipulated image is
\[
x^{(m)}=\mathcal{T}_m(x).
\]
These manipulations alter visible appearance cues but do not change the subject's true age.
A robust age verification model should therefore preserve the binary decision under such changes:
\[
h_f(x;\tau)=h_f(x^{(m)};\tau).
\]

\subsection{Datasets}
\label{sec:datasets}

In the experimental analysis, we consider three widely used face datasets. 

\textbf{UTKFace}~\cite{zhang2017age} contains approximately 24,000 face images
with exact age labels ranging from 0 to 116. It also provides gender
annotations and self-reported race labels over five categories: White, Black,
Asian, Indian, and Other. 

\textbf{Adience}~\cite{eidinger2014age} contains 26,580 face images
annotated with age intervals rather than exact ages. The available intervals are 0--2, 4--6, 8--13, 15--20, 25--32, 38--43,
48--53, and 60--100.

\textbf{FairFace}~\cite{karkkainen2021fairface} includes more than 100,000 face images annotated with
gender, race, and coarse age groups. The age groups are 0--2, 3--9, 10--19,
20--29, 30--39, 40--49, 50--59, 60--69, and  70+. 

\subsection{Manipulations}
\label{sec:manipulations}
\begin{figure}[t]
\centering

\begin{subfigure}{\linewidth}
    \centering
    \includegraphics[width=\linewidth]{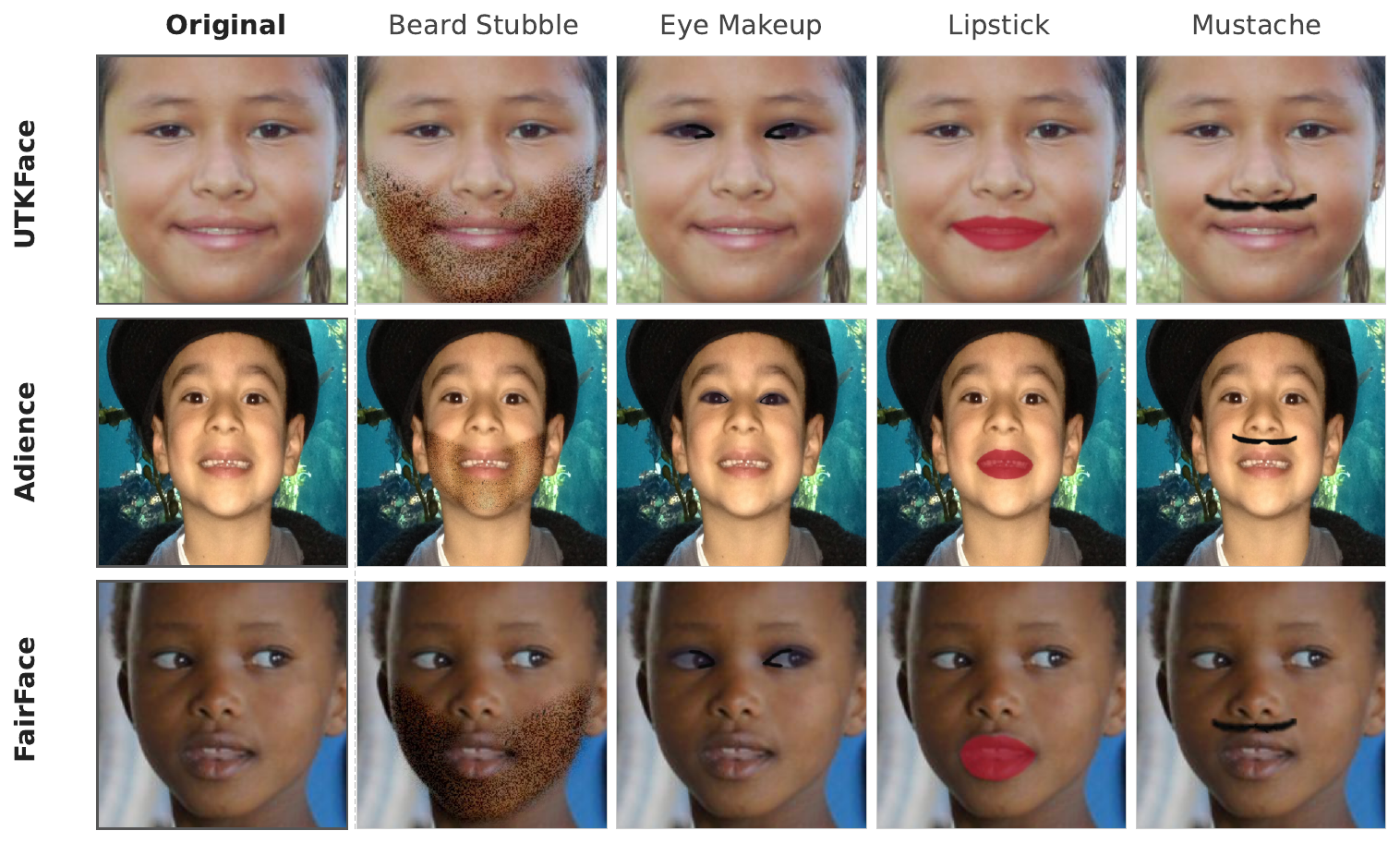}
    \caption{OpenCV manipulations.}
    \label{fig:showcase}
\end{subfigure}

\vspace{0.5em}

\begin{subfigure}{\linewidth}
    \centering
    \includegraphics[width=\linewidth]{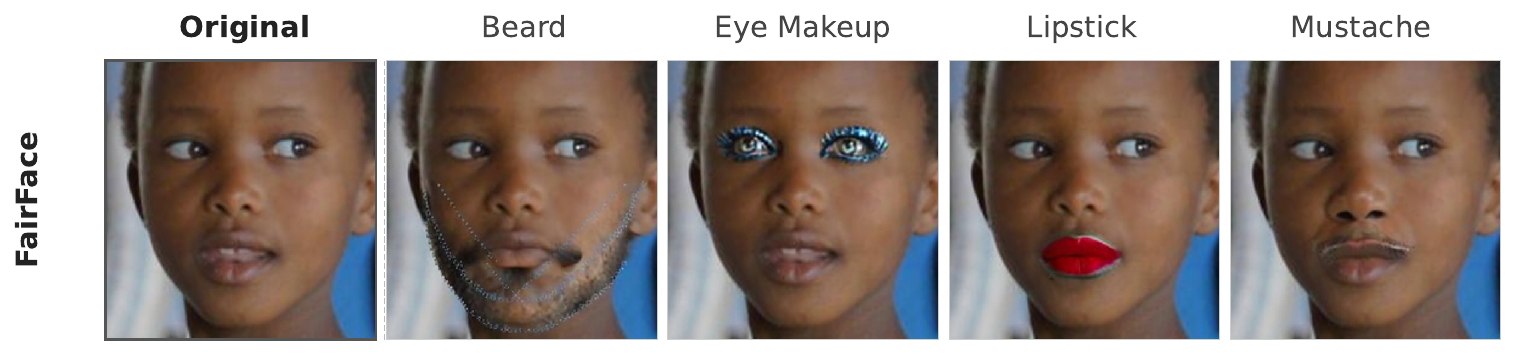}
    \caption{Generative manipulations.}
    \label{fig:example_genai}
\end{subfigure}

\caption{Examples of the visual manipulations considered in this study.}
\label{fig:examples}

\end{figure}

Each manipulation is implemented as a deterministic OpenCV\footnote{\url{https://opencv.org/}} overlay aligned
with facial landmarks extracted using MediaPipe FaceMesh~\cite{kartynnik2019}.
This design is deliberately simplistic as we target manipulations that can approximate what can be produced with a marker or common cosmetics, rather than creating optimized or photorealistic attacks. 
Figure~\ref{fig:showcase} shows examples of all four manipulations, i.e., mustache, beard stubble, eye makeup, and lipstick. Since the overlays are deterministic and landmark-aligned, the paired original and manipulated images differ only in the applied appearance cue. This allows the evaluation to isolate the effect of the manipulation from subject-level variation. However, for completeness, we consider an ablation using Stable Diffusion 2 inpainting \cite{rombach2022sdimpainting} to generate photorealistic manipulated images, as shown in Figure~\ref{fig:example_genai}.

\subsection{Evaluation Sets}
\label{sec:testsets}

For each dataset, we construct two evaluation sets, the Original Set (Set~A) and the Robustness Set (Set~B).
Set~A serves as a baseline to measure whether each model performs reasonably well on original images,
while Set~B is used to evaluate how the appearance manipulations can affect the models' behavior.

\textbf{Original Set (Set~A)} is a balanced evaluation set with
1,000 samples per dataset: 500 underage and 500 age-eligible samples. For UTKFace, underage samples
are derived from ages 6--12 and age-eligible samples from ages 14--20. For Adience,
underage samples are derived from the age intervals 4--6 and
8--13, while age-eligible instances are sampled from the 15--20 interval. For FairFace, underage images are sampled from the 3--9 age group and age-eligible samples from the 20--29 age group. We avoid the FairFace 10--19 group in Set~A because it crosses the 13-year verification threshold.

\textbf{Robustness Set (Set~B)} contains only underage subjects, each represented by the original image and its manipulated variants.
For UTKFace, we use samples from ages 6--12. For Adience, we consider the samples from the available intervals used for the underage pool. For FairFace, we use the 3--9 age group, and all samples are drawn from the official test set because one of the tested models was trained on FairFace and because we reserve the training split for the mitigation experiments in Section~\ref{sec:mitigation}. Each image in Set~B appears once in its original form and four times manipulated, once per manipulation type. This setting allows for the isolation of the effect of each manipulation condition.

\subsection{Models}
\label{sec:models}

We evaluate seven models grouped into three families: vision models,
VLMs, and MLLMs.

\textbf{Vision models.}
FairFace~\cite{karkkainen2021fairface}
uses a ResNet-34 backbone trained for face attribute prediction, including age, gender, and race.
MiVOLO~\cite{kuprashevich2023mivolo} is a transformer-based age and gender estimation
model utilizing both facial and body features. We evaluate it in face-only mode as the datasets used in this study provide face crops rather than full person crops.  
The ViT-Age classifier~\cite{nateraw_vit} uses a ViT-B/16 backbone trained for age classification. For classification-based vision models, following prior work~\cite{rothe2015dex}, we convert the output distribution into a scalar age estimate using the expected value over age classes.

\textbf{VLMs.}
This group includes the fine-tuned SigLIP2~\cite{vlms} and MetaCLIP2~\cite{metaclip2} for predicting age groups. 

\textbf{MLLMs.}
This group includes instruction-following multimodal models that take an image and a natural-language prompt as input and return a textual answer in a zero-shot manner. We evaluate Qwen2.5-VL~\cite{bai2023qwen} and LLaVA-1.6~\cite{liu2024llava} in this setting. For each image, the model is prompted to estimate the subject's age, and the generated response is parsed into a scalar value. In particular, we use the following prompt: ``\textit{How old does the person in this image appear to be? Reply with only a single integer number representing the age in years. Do not include any other text, units, or explanation}''.

\subsection{Evaluation Protocol}
\label{sec:metrics}

Overall performance is reported using mean absolute error (MAE) on Set~A. For the age verification setting, we convert each scalar age estimate into a binary decision at \(\tau=13\), where predictions below the threshold are treated as underage. On Set~B, accuracy measures the fraction of underage subjects that are correctly classified, i.e., True Negative Rate (TNR). In addition, for each manipulation, we consider the flip rate, i.e., the fraction of underage samples that are correctly classified before manipulation but misclassified as age-eligible afterwards.
This metric targets the age verification bypass case directly. We also report the mean predicted age shift, computed as the average difference between the manipulated and original age predictions on Set~B.
All Set~B metrics are reported with 95\% bootstrap confidence intervals using 1,000 bootstrap resamples.
\begin{table}[h]
  \centering
  \caption{MAE performance (in years) on Set A. Lower is better; the best result in each column is shown in bold.}
  \label{tab:set_a_mae}
  \begin{tabular}{lcccc}
    \toprule
    Model & UTKFace & Adience & FairFace & Average \\
    \midrule
FairFace    & 3.2 & 4.0 & \textbf{3.0} & 3.4 \\
MiVOLO      & 3.8 & 6.0 & 5.5 & 5.1 \\
ViT-Age     & 3.3 & 3.4 & 3.2 & 3.3 \\
MetaCLIP2   & 8.8 & 9.7 & 8.6 & 9.0 \\
SigLIP2     & 8.1 & 7.9 & 6.1 & 7.3 \\
Qwen2.5-VL  & 3.3 & 5.9 & 4.0 & 4.4 \\
LLaVA-1.6   & \textbf{3.0} & \textbf{2.9} & 3.6 & \textbf{3.2} \\
    \bottomrule
  \end{tabular}
\end{table}
\begin{table}[H]
  \centering
  \scriptsize
  \setlength{\tabcolsep}{4pt}
  \renewcommand{\arraystretch}{1.1}
  \caption{Set B TNR (\%) before and after manipulation. Parentheses show the change compared to the original; changes of at least 3\% in magnitude are colored \textcolor{red}{red} for decreases and \textcolor{blue}{blue} for increases, while smaller changes are shown in \textcolor{gray}{gray}.}
  \label{tab:binary_threshold_all}
  \resizebox{\textwidth}{!}{%
  \begin{tabular}{llllll}
    \toprule
    Model & Original & Beard Stubble & Eye Makeup & Lipstick & Mustache \\
    \midrule
    \multicolumn{6}{l}{\textbf{UTKFace}} \\
    \midrule
    FairFace & $83.7_{\pm 2.1}$ & $68.7_{\pm 2.8}$ $(\textcolor{red}{-15.0})$ & $81.7_{\pm 2.2}$ $(\textcolor{gray}{-2.0})$ & $79.2_{\pm 2.5}$ $(\textcolor{red}{-4.5})$ & $84.0_{\pm 2.2}$ $(\textcolor{gray}{+0.3})$ \\
        MiVOLO & $80.4_{\pm 2.4}$ & $31.3_{\pm 2.9}$ $(\textcolor{red}{-49.0})$ & $65.0_{\pm 2.8}$ $(\textcolor{red}{-15.3})$ & $70.5_{\pm 2.6}$ $(\textcolor{red}{-9.9})$ & $86.7_{\pm 2.0}$ $(\textcolor{blue}{+6.3})$ \\
    ViT-Age & $84.9_{\pm 2.1}$ & $53.3_{\pm 2.9}$ $(\textcolor{red}{-31.5})$ & $75.5_{\pm 2.4}$ $(\textcolor{red}{-9.3})$ & $81.8_{\pm 2.3}$ $(\textcolor{red}{-3.0})$ & $78.7_{\pm 2.5}$ $(\textcolor{red}{-6.2})$ \\
    SigLIP2 & $70.7_{\pm 2.8}$ & $57.2_{\pm 2.9}$ $(\textcolor{red}{-13.4})$ & $69.9_{\pm 2.8}$ $(\textcolor{gray}{-0.8})$ & $42.8_{\pm 2.9}$ $(\textcolor{red}{-27.9})$ & $68.9_{\pm 2.8}$ $(\textcolor{gray}{-1.8})$ \\
    MetaCLIP2 & $67.7_{\pm 2.9}$ & $55.2_{\pm 3.0}$ $(\textcolor{red}{-12.5})$ & $74.0_{\pm 2.7}$ $(\textcolor{blue}{+6.3})$ & $58.2_{\pm 3.0}$ $(\textcolor{red}{-9.5})$ & $67.5_{\pm 2.9}$ $(\textcolor{gray}{-0.2})$ \\
    Qwen2.5-VL & $96.2_{\pm 1.2}$ & $80.3_{\pm 2.3}$ $(\textcolor{red}{-15.9})$ & $92.8_{\pm 1.6}$ $(\textcolor{red}{-3.4})$ & $86.0_{\pm 2.2}$ $(\textcolor{red}{-10.2})$ & $93.7_{\pm 1.5}$ $(\textcolor{gray}{-2.5})$ \\
        LLaVA-1.6 & $98.0_{\pm 0.8}$ & $91.0_{\pm 1.8}$ $(\textcolor{red}{-7.0})$ & $97.1_{\pm 1.0}$ $(\textcolor{gray}{-0.9})$ & $94.7_{\pm 1.4}$ $(\textcolor{red}{-3.3})$ & $98.5_{\pm 0.7}$ $(\textcolor{gray}{+0.5})$ \\

    \addlinespace[4pt]
    \multicolumn{6}{l}{\textbf{Adience}} \\
    \midrule
    FairFace & $93.6_{\pm 1.0}$ & $65.3_{\pm 1.9}$ $(\textcolor{red}{-28.3})$ & $94.0_{\pm 1.0}$ $(\textcolor{gray}{+0.4})$ & $90.3_{\pm 1.3}$ $(\textcolor{red}{-3.3})$ & $91.2_{\pm 1.2}$ $(\textcolor{gray}{-2.4})$ \\
    
        MiVOLO & $92.0_{\pm 1.2}$ & $53.0_{\pm 2.1}$ $(\textcolor{red}{-39.0})$ & $85.1_{\pm 1.5}$ $(\textcolor{red}{-6.8})$ & $85.5_{\pm 1.4}$ $(\textcolor{red}{-6.4})$ & $92.6_{\pm 1.1}$ $(\textcolor{gray}{+0.6})$ \\
    ViT-Age & $98.7_{\pm 0.5}$ & $88.7_{\pm 1.4}$ $(\textcolor{red}{-9.9})$ & $98.0_{\pm 0.6}$ $(\textcolor{gray}{-0.6})$ & $98.2_{\pm 0.6}$ $(\textcolor{gray}{-0.5})$ & $97.7_{\pm 0.6}$ $(\textcolor{gray}{-1.0})$ \\
    SigLIP2 & $91.7_{\pm 1.1}$ & $77.0_{\pm 1.8}$ $(\textcolor{red}{-14.8})$ & $92.1_{\pm 1.2}$ $(\textcolor{gray}{+0.4})$ & $75.5_{\pm 1.9}$ $(\textcolor{red}{-16.2})$ & $92.2_{\pm 1.1}$ $(\textcolor{gray}{+0.4})$ \\
    MetaCLIP2 & $78.1_{\pm 1.8}$ & $69.0_{\pm 1.9}$ $(\textcolor{red}{-9.1})$ & $82.2_{\pm 1.7}$ $(\textcolor{blue}{+4.2})$ & $72.8_{\pm 1.9}$ $(\textcolor{red}{-5.3})$ & $79.4_{\pm 1.7}$ $(\textcolor{gray}{+1.3})$ \\

    Qwen2.5-VL & $93.8_{\pm 1.0}$ & $87.5_{\pm 1.4}$ $(\textcolor{red}{-6.4})$ & $93.6_{\pm 1.1}$ $(\textcolor{gray}{-0.2})$ & $87.7_{\pm 1.3}$ $(\textcolor{red}{-6.2})$ & $92.3_{\pm 1.1}$ $(\textcolor{gray}{-1.5})$ \\
        LLaVA-1.6 & $99.8_{\pm 0.2}$ & $99.6_{\pm 0.3}$ $(\textcolor{gray}{-0.2})$ & $99.7_{\pm 0.3}$ $(\textcolor{gray}{-0.1})$ & $99.7_{\pm 0.3}$ $(\textcolor{gray}{-0.1})$ & $99.8_{\pm 0.2}$ $(\textcolor{gray}{0.0})$ \\

    \addlinespace[4pt]
    \multicolumn{6}{l}{\textbf{FairFace}} \\
    \midrule
    FairFace & $95.7_{\pm 1.1}$ & $88.4_{\pm 1.8}$ $(\textcolor{red}{-7.2})$ & $96.9_{\pm 0.9}$ $(\textcolor{gray}{+1.2})$ & $92.3_{\pm 1.5}$ $(\textcolor{red}{-3.4})$ & $95.1_{\pm 1.2}$ $(\textcolor{gray}{-0.6})$ \\
    
        MiVOLO & $77.3_{\pm 2.3}$ & $29.8_{\pm 2.5}$ $(\textcolor{red}{-47.5})$ & $68.5_{\pm 2.6}$ $(\textcolor{red}{-8.8})$ & $68.8_{\pm 2.6}$ $(\textcolor{red}{-8.4})$ & $81.7_{\pm 2.2}$ $(\textcolor{blue}{+4.4})$ \\
    ViT-Age & $94.1_{\pm 1.3}$ & $73.0_{\pm 2.5}$ $(\textcolor{red}{-21.1})$ & $94.9_{\pm 1.2}$ $(\textcolor{gray}{+0.7})$ & $93.7_{\pm 1.4}$ $(\textcolor{gray}{-0.5})$ & $92.7_{\pm 1.4}$ $(\textcolor{gray}{-1.4})$ \\
    SigLIP2 & $84.1_{\pm 2.0}$ & $76.7_{\pm 2.4}$ $(\textcolor{red}{-7.4})$ & $87.0_{\pm 1.9}$ $(\textcolor{gray}{+2.9})$ & $61.6_{\pm 2.7}$ $(\textcolor{red}{-22.5})$ & $84.2_{\pm 1.9}$ $(\textcolor{gray}{+0.1})$ \\
    MetaCLIP2 & $71.2_{\pm 2.4}$ & $64.6_{\pm 2.7}$ $(\textcolor{red}{-6.7})$ & $80.0_{\pm 2.2}$ $(\textcolor{blue}{+8.8})$ & $67.8_{\pm 2.6}$ $(\textcolor{red}{-3.5})$ & $73.7_{\pm 2.4}$ $(\textcolor{gray}{+2.4})$ \\

    Qwen2.5-VL & $96.4_{\pm 1.1}$ & $87.9_{\pm 1.8}$ $(\textcolor{red}{-8.5})$ & $95.7_{\pm 1.1}$ $(\textcolor{gray}{-0.6})$ & $89.9_{\pm 1.8}$ $(\textcolor{red}{-6.5})$ & $95.5_{\pm 1.1}$ $(\textcolor{gray}{-0.9})$ \\
        LLaVA-1.6 & $98.9_{\pm 0.6}$ & $97.5_{\pm 0.8}$ $(\textcolor{gray}{-1.4})$ & $99.0_{\pm 0.5}$ $(\textcolor{gray}{+0.2})$ & $98.1_{\pm 0.7}$ $(\textcolor{gray}{-0.8})$ & $99.0_{\pm 0.6}$ $(\textcolor{gray}{+0.2})$ \\

    \midrule
    Average & $87.9_{\pm 1.5}$ & $71.2_{\pm 2.1}$ $(\textcolor{red}{-16.8})$ & $86.8_{\pm 1.6}$ $(\textcolor{gray}{-1.1})$ & $80.7_{\pm 1.8}$ $(\textcolor{red}{-7.2})$ & $87.9_{\pm 1.5}$ $(\textcolor{gray}{-0.1})$ \\
    \bottomrule
  \end{tabular}
  }
\end{table}

\section{Results}
\label{sec:results}

\subsection{Overall Performance}
\label{sec:baseline}

As shown in Table~\ref{tab:set_a_mae}, most of the evaluated models demonstrate reasonable age estimation performance on the original images (Set A), establishing a credible baseline for the subsequent robustness analysis. LLaVA-1.6 achieves the lowest average MAE across datasets at 3.2 years, followed closely by ViT-Age and FairFace at 3.3 and 3.4 years, respectively. Performance remains competitive for Qwen2.5-VL and MiVOLO, although their performance varies more substantially across datasets. The VLMs, SigLIP2 and MetaCLIP2, exhibit higher MAEs, but their inclusion in our analysis allows us to examine whether models with weaker overall age estimates are also more sensitive to manipulations.

\subsection{Robustness to Manipulations}
\label{sec:results_manip}

\begin{table}[t]
  \centering
  \scriptsize
  \setlength{\tabcolsep}{4pt}
  \caption{Flip rate (\%) under each manipulation. The highest flip rate for each dataset and manipulation is shown in bold. }
  \label{tab:flip_rate_all}
  \begin{tabular}{lcccc}
    \toprule
    Model & Beard Stubble & Eye Makeup & Lipstick & Mustache \\
    \midrule
    \multicolumn{5}{l}{\textbf{UTKFace}} \\
    \midrule
    FairFace & $19.0_{\pm 2.6}$ & $7.4_{\pm 1.6}$ & $6.9_{\pm 1.7}$ & $4.9_{\pm 1.4}$ \\
        MiVOLO & $\mathbf{61.0_{\pm 3.5}}$ & $\mathbf{22.4_{\pm 2.6}}$ & $13.2_{\pm 2.2}$ & $0.6_{\pm 0.5}$ \\
    ViT-Age & $37.3_{\pm 3.2}$ & $12.8_{\pm 2.1}$ & $6.5_{\pm 1.5}$ & $\mathbf{8.4_{\pm 1.8}}$ \\
    SigLIP2 & $21.0_{\pm 2.8}$ & $7.8_{\pm 2.0}$ & $\mathbf{39.5_{\pm 3.4}}$ & $5.4_{\pm 1.7}$ \\
    MetaCLIP2 & $23.6_{\pm 3.0}$ & $3.7_{\pm 1.3}$ & $17.7_{\pm 2.7}$ & $4.6_{\pm 1.5}$ \\

    Qwen2.5-VL & $16.5_{\pm 2.2}$ & $4.2_{\pm 1.2}$ & $10.7_{\pm 1.9}$ & $3.1_{\pm 1.0}$ \\
        LLaVA-1.6 & $7.3_{\pm 1.6}$ & $1.3_{\pm 0.7}$ & $3.4_{\pm 1.1}$ & $0.1_{\pm 0.1}$ \\

    \addlinespace[4pt]
    \multicolumn{5}{l}{\textbf{Adience}} \\
    \midrule
    FairFace & $30.3_{\pm 1.8}$ & $1.1_{\pm 0.5}$ & $4.7_{\pm 0.9}$ & $\mathbf{3.5_{\pm 0.8}}$ \\
        MiVOLO & $\mathbf{42.5_{\pm 2.2}}$ & $\mathbf{8.7_{\pm 1.3}}$ & $7.5_{\pm 1.1}$ & $0.9_{\pm 0.4}$ \\
    ViT-Age & $10.1_{\pm 1.2}$ & $0.8_{\pm 0.4}$ & $0.8_{\pm 0.4}$ & $1.1_{\pm 0.5}$ \\
    SigLIP2 & $16.2_{\pm 1.6}$ & $2.1_{\pm 0.7}$ & $\mathbf{17.8_{\pm 1.7}}$ & $1.3_{\pm 0.5}$ \\
    MetaCLIP2 & $14.4_{\pm 1.7}$ & $1.7_{\pm 0.6}$ & $9.7_{\pm 1.4}$ & $2.6_{\pm 0.8}$ \\

    Qwen2.5-VL & $7.4_{\pm 1.2}$ & $2.0_{\pm 0.6}$ & $7.1_{\pm 1.1}$ & $2.4_{\pm 0.7}$ \\
        LLaVA-1.6 & $0.2_{\pm 0.2}$ & $0.1_{\pm 0.1}$ & $0.1_{\pm 0.1}$ & $0.0_{\pm 0.0}$ \\

    \addlinespace[4pt]
    \multicolumn{5}{l}{\textbf{FairFace}} \\
    \midrule
    FairFace & $7.6_{\pm 1.4}$ & $1.1_{\pm 0.6}$ & $3.9_{\pm 1.0}$ & $1.8_{\pm 0.7}$ \\
        MiVOLO & $\mathbf{61.4_{\pm 3.1}}$ & $\mathbf{14.4_{\pm 2.2}}$ & $11.9_{\pm 2.2}$ & $1.5_{\pm 0.8}$ \\
        ViT-Age & $22.5_{\pm 2.3}$ & $1.5_{\pm 0.7}$ & $2.5_{\pm 0.9}$ & $2.6_{\pm 1.0}$ \\
SigLIP2 & $11.3_{\pm 2.0}$ & $2.6_{\pm 1.0}$ & $\mathbf{26.7_{\pm 2.6}}$ & $2.8_{\pm 1.0}$ \\
    MetaCLIP2 & $16.0_{\pm 2.4}$ & $3.3_{\pm 1.1}$ & $9.7_{\pm 2.0}$ & $\mathbf{3.5_{\pm 1.2}}$ \\

    Qwen2.5-VL & $9.3_{\pm 1.6}$ & $1.8_{\pm 0.7}$ & $7.1_{\pm 1.4}$ & $1.5_{\pm 0.7}$ \\
        LLaVA-1.6 & $1.4_{\pm 0.6}$ & $0.1_{\pm 0.1}$ & $0.8_{\pm 0.5}$ & $0.1_{\pm 0.1}$ \\

    \midrule
     Average & $20.8_{\pm 2.0}$ & $4.8_{\pm 1.1}$ & $9.9_{\pm 1.5}$ & $2.5_{\pm 0.8}$ \\
    \bottomrule
  \end{tabular}
\end{table}

\begin{table}[t]
  \centering
  \scriptsize
  \setlength{\tabcolsep}{6pt}
  \renewcommand{\arraystretch}{1.1}
  \caption{Mean change in predicted age (years) after manipulation relative to the original images. Shifts above $+1$ years are colored \textcolor{red}{red} while shifts below $-1$ years are colored \textcolor{blue}{blue}.}
  \label{tab:delta_predictions_all}
  \begin{tabular}{lcccc}
    \toprule
    Model & Beard Stubble & Eye Makeup & Lipstick & Mustache \\
    \midrule
    \multicolumn{5}{l}{\textbf{UTKFace}} \\
    \midrule
    FairFace & $\textcolor{red}{+1.8_{\pm 0.2}}$ & $-0.1_{\pm 0.1}$ & $+0.6_{\pm 0.1}$ & $+0.2_{\pm 0.1}$ \\
        MiVOLO & $\textcolor{red}{+7.1_{\pm 0.4}}$ & $\textcolor{red}{+1.5_{\pm 0.2}}$ & $\textcolor{red}{+1.1_{\pm 0.1}}$ & $-0.4_{\pm 0.1}$ \\
    ViT-Age & $\textcolor{red}{+4.1_{\pm 0.2}}$ & $+0.8_{\pm 0.2}$ & $+0.3_{\pm 0.2}$ & $+1.0_{\pm 0.1}$ \\
    MetaCLIP2 & $\textcolor{red}{+1.5_{\pm 0.6}}$ & $\textcolor{blue}{-1.2_{\pm 0.5}}$ & $\textcolor{red}{+4.0_{\pm 0.6}}$ & $-0.8_{\pm 0.4}$ \\
    SigLIP2 & $\textcolor{red}{+2.0_{\pm 0.4}}$ & $+0.1_{\pm 0.4}$ & $\textcolor{red}{+8.6_{\pm 0.7}}$ & $+0.2_{\pm 0.3}$ \\

    Qwen2.5-VL & $\textcolor{red}{+3.3_{\pm 0.3}}$ & $-0.4_{\pm 0.2}$ & $\textcolor{red}{+1.3_{\pm 0.3}}$ & $+0.4_{\pm 0.1}$ \\
        LLaVA-1.6 & $\textcolor{red}{+1.5_{\pm 0.2}}$ & $-0.1_{\pm 0.1}$ & $+0.2_{\pm 0.1}$ & $+0.2_{\pm 0.1}$ \\
\addlinespace[4pt]
    \multicolumn{5}{l}{\textbf{Adience}} \\
    \midrule
    FairFace & $\textcolor{red}{+5.6_{\pm 0.3}}$ & $-0.7_{\pm 0.1}$ & $+0.6_{\pm 0.1}$ & $+0.6_{\pm 0.1}$ \\
        MiVOLO & $\textcolor{red}{+6.8_{\pm 0.3}}$ & $+0.6_{\pm 0.1}$ & $+0.8_{\pm 0.1}$ & $0.0_{\pm 0.0}$ \\
    ViT-Age & $\textcolor{red}{+2.3_{\pm 0.2}}$ & $-0.4_{\pm 0.1}$ & $+0.2_{\pm 0.1}$ & $+1.0_{\pm 0.0}$ \\
    SigLIP2 & $\textcolor{red}{+2.8_{\pm 0.3}}$ & $-0.1_{\pm 0.2}$ & $\textcolor{red}{+4.1_{\pm 0.4}}$ & $-0.2_{\pm 0.2}$ \\
    MetaCLIP2 & $\textcolor{red}{+1.8_{\pm 0.4}}$ & $-0.8_{\pm 0.3}$ & $\textcolor{red}{+1.7_{\pm 0.3}}$ & $-0.8_{\pm 0.2}$ \\
    Qwen2.5-VL & $\textcolor{red}{+1.8_{\pm 0.2}}$ & $-0.7_{\pm 0.2}$ & $\textcolor{red}{+1.7_{\pm 0.3}}$ & $+0.5_{\pm 0.1}$ \\
    LLaVA-1.6 & $+0.2_{\pm 0.1}$ & $-0.2_{\pm 0.1}$ & $-0.1_{\pm 0.0}$ & $+0.1_{\pm 0.0}$ \\

    \addlinespace[4pt]
    \multicolumn{5}{l}{\textbf{FairFace}} \\
    \midrule
    FairFace & $\textcolor{red}{+1.3_{\pm 0.1}}$ & $-0.5_{\pm 0.1}$ & $+0.5_{\pm 0.1}$ & $+0.4_{\pm 0.1}$ \\
        MiVOLO & $\textcolor{red}{+8.6_{\pm 0.4}}$ & $+1.0_{\pm 0.2}$ & $\textcolor{red}{+1.2_{\pm 0.2}}$ & $-0.4_{\pm 0.1}$ \\
    ViT-Age & $\textcolor{red}{+3.3_{\pm 0.2}}$ & $-0.5_{\pm 0.1}$ & $+0.1_{\pm 0.1}$ & $+0.7_{\pm 0.1}$ \\
    SigLIP2 & $\textcolor{red}{+1.2_{\pm 0.3}}$ & $-0.6_{\pm 0.3}$ & $\textcolor{red}{+4.8_{\pm 0.5}}$ & $-0.3_{\pm 0.2}$ \\
    MetaCLIP2 & $+0.5_{\pm 0.6}$ & $\textcolor{blue}{-1.8_{\pm 0.5}}$ & $+1.0_{\pm 0.5}$ & $\textcolor{blue}{-1.3_{\pm 0.4}}$ \\

    Qwen2.5-VL & $\textcolor{red}{+2.0_{\pm 0.2}}$ & $-0.8_{\pm 0.2}$ & $+0.8_{\pm 0.2}$ & $+0.3_{\pm 0.1}$ \\
        LLaVA-1.6 & $+0.7_{\pm 0.1}$ & $-0.4_{\pm 0.1}$ & $0.0_{\pm 0.1}$ & $+0.4_{\pm 0.1}$ \\

    \midrule
    Average & $\textcolor{red}{+2.9_{\pm 0.3}}$ & $-0.2_{\pm 0.2}$ & $\textcolor{red}{+1.6_{\pm 0.2}}$ & $+0.1_{\pm 0.1}$ \\
    \bottomrule
  \end{tabular}
  
\end{table}

\begin{table*}[h]
  \centering
  \scriptsize
  \setlength{\tabcolsep}{4pt}
  \renewcommand{\arraystretch}{1.1}
\caption{Comparison of OpenCV-based and generative facial manipulations, macro-averaged across all models and datasets. We report TNR, flip rate, and mean change in predicted age.}
  \label{tab:genai}
\begin{tabular}{lcccc} \toprule Metric & Beard & Eye Makeup & Lipstick & Mustache \\ \midrule TNR (OpenCV) & $71.2_{\pm 2.1}$ & $86.8_{\pm 1.6}$ & $80.7_{\pm 1.8}$ & $87.9_{\pm 1.5}$ \\ TNR (Generative) & $57.1_{\pm 2.3}$ & $81.9_{\pm 1.8}$  & $81.9_{\pm 1.8}$ & $75.8_{\pm 2.1}$\\ \midrule Flip Rate (OpenCV) & $20.8_{\pm 2.0}$ & $4.8_{\pm 1.1}$ & $9.9_{\pm 1.5}$ & $2.5_{\pm 0.8}$ \\ Flip Rate (Generative) & $36.3_{\pm 2.5}$ & $9.8_{\pm 1.5}$ & $8.8_{\pm 1.4}$ & $15.7_{\pm 1.9}$ \\ \midrule $\Delta$ Mean pred. age (OpenCV) & $+2.9_{\pm 0.3}$ & $-0.2_{\pm 0.2}$ & $+1.6_{\pm 0.2}$ & $+0.1_{\pm 0.1}$ \\ $\Delta$ Mean pred. age (Generative) & $+6.5_{\pm 0.5}$ & $+0.8_{\pm 0.2}$ & $+1.3_{\pm 0.2}$ & $+2.3_{\pm 0.3}$ \\ \bottomrule \end{tabular}
\end{table*}

Table~\ref{tab:binary_threshold_all} reports TNR on Set B before and after manipulation. Across all models and datasets, beard stubble causes the largest average TNR reduction, decreasing TNR by 16.8 percentage points, from 87.9\% to 71.2\%. Lipstick produces the second-largest average reduction of 7.2\%, whereas eye makeup and mustache have substantially smaller aggregate effects of 1.1\% and 0.1\%, respectively. The severity of these effects varies considerably across models. MiVOLO is particularly vulnerable to beard stubble, with TNR reductions of 49.0\%, 39.0\%, and 47.5\% on UTKFace, Adience, and FairFace, respectively. SigLIP2 is especially sensitive to lipstick, losing between 16.2 and 27.9 percentage points across datasets. In contrast, LLaVA-1.6 remains consistently robust, with post-manipulation TNR generally remaining above 91\%. This robustness can stem from several factors, including its training data, architecture, or other bias-related considerations and objectives. 

The flip rates in Table~\ref{tab:flip_rate_all} confirm that these errors directly translate to severe age verification bypassing issues. Beard stubble produces the highest average flip rate at 20.8\%, followed by lipstick at 9.9\%, eye makeup at 4.8\%, and mustache at 2.5\%. MiVOLO exhibits the highest beard-stubble flip rate on every dataset, ranging from 42.5\% on Adience to 61.4\% on FairFace. Similar to the previous findings, LLaVA-1.6 consistently achieves the lowest flip rates, remaining below 1.5\% for nearly all dataset--manipulation combinations except beard stubble and lipstick on UTKFace, which is overall the most challenging benchmark.
Finally, Table~\ref{tab:delta_predictions_all} shows that overall, beard stubble and lipstick shift the models' age predictions by 2.9 and 1.6 years on average, respectively.

\subsection{Robustness to Generative Manipulations}

As an ablation, we evaluate photorealistic manipulations produced using Stable Diffusion 2 inpainting (Figure~\ref{fig:example_genai}). Table~\ref{tab:genai} shows that generative manipulations generally affect the evaluated models more severely than the simple OpenCV overlays. For example, generative beards increase the average flip rate from 20.8\% to 36.3\% and increase the mean predicted age by 6.5 years, compared with 2.9 years for the OpenCV version. Generative mustaches similarly increase the flip rate from 2.5\% to 15.7\%. These findings demonstrate that more realistic alterations can amplify model vulnerability. However, the central finding does not depend on access to generative tools or photorealistic edits. Even visibly artificial, simulation of cues that an underage individual could draw on their face are sufficient to bypass age verification systems.

\subsection{Demographic Disparities}
\label{sec:demographics}

The model-averaged results reported in Table~\ref{tab:demographic_race_average_fairface} reveal that vulnerability varies considerably across racial groups. Indian subjects exhibit the highest average flip rates under beard stubble and eye makeup, at 24\% and 5\%, respectively, while Black subjects are most affected by lipstick at 11.4\%. Middle Eastern subjects have the highest mustache flip rate at 2.8\%. Overall, Indian, Black, and Middle Eastern samples show consistently higher flip rates than other races.  In contrast, East Asian subjects consistently exhibit among the lowest flip rates across all manipulations.
Moreover, as shown in Table~\ref{tab:demographic_gender_average_fairface}, female subjects exhibit higher model-averaged flip rates than male subjects for all manipulations. The largest gaps occur for the female-related manipulations, namely eye makeup, where the female flip rate is approximately twice the male rate (4.8\% versus 2.4\%), and lipstick (10.2\% versus 7.8\%).


\begin{table}[t]
\centering
\small
\setlength{\tabcolsep}{4pt}
  \renewcommand{\arraystretch}{1.1}
\caption{Flip rate (\%) by race group on FairFace, averaged across models.}
\label{tab:demographic_race_average_fairface}
\begin{tabular}{lrrrr}
\toprule
Race & \multicolumn{1}{c}{Beard Stubble} & \multicolumn{1}{c}{Eye Makeup} & \multicolumn{1}{c}{Lipstick} & \multicolumn{1}{c}{Mustache} \\
\midrule
White & 17.7 & 3.3 & 10.41 & 2.4 \\
Black & 23.7 & 4.5 & \textbf{11.4} & 2.5 \\
East Asian & 13.9 & 1.8 & 6.1 & 0.8 \\
SE Asian & 15.2 & 3.2 & 6.4 & 1.9 \\
Indian & \textbf{24.0} & \textbf{5.0} & 10.1 & 2.2 \\
Middle Eastern & 19.2 & 4.4 & 10.6 & \textbf{2.8} \\
Latino & 17.9 & 3.8 & 9.2 & 1.8 \\
\bottomrule
\end{tabular}
\end{table}

\begin{table}[t]
\centering
\small
\setlength{\tabcolsep}{4pt}
  \renewcommand{\arraystretch}{1.1}
\caption{Flip rate (\%) by gender on FairFace, averaged across models.}
\label{tab:demographic_gender_average_fairface}
\begin{tabular}{lrrrr}
\toprule
Gender & \multicolumn{1}{c}{Beard Stubble} & \multicolumn{1}{c}{Eye Makeup} & \multicolumn{1}{c}{Lipstick} & \multicolumn{1}{c}{Mustache} \\
\midrule
Male & 17.9 & 2.4 & 7.8 & 1.9 \\
Female & \textbf{19.2} & \textbf{4.8} & \textbf{10.2} & \textbf{2.0} \\
\bottomrule
\end{tabular}
\end{table}

\section{Mitigation}
\label{sec:mitigation}

The preceding analysis shows that facial manipulations can cause large and systematic failures across age verification models. 
We therefore investigate whether robustness can be improved without retraining the underlying backbone. Specifically, we train a lightweight two-layer MLP classifier on the frozen extracted features to evaluate whether performance can be recovered under manipulation-induced shifts.
Note that we focus on frozen-backbone linear probing rather than end-to-end fine-tuning to avoid compatibility issues between bias mitigation approaches and certain backbones (i.e., MLLMs).

\subsection{Setup}
\label{sec:mit_setup}

\textbf{Protocol.}
\textit{Underage} and \textit{age-eligible} are the two target classes. For training, we use the official FairFace training set. For underage samples, we additionally include their manipulated variants, using the deterministic OpenCV-based overlays defined in Section~\ref{sec:manipulations}.  We vary the proportion of manipulated training examples relative to the total number of underage samples across $\{1\%, 5\%, 10\%\}$ to evaluate robustness under different levels of distribution shift. 
The Set B is used as a test set. The minimum TNR across all four manipulation groups, i.e., Worst Group TNR (\wga{}), and the flip rate are considered as evaluation metrics.
Results are reported as the mean and standard deviation over three random seeds.

\textbf{Methods.}
We focus on five representative approaches: Empirical Risk Minimization
(ERM), which serves as a baseline; LfF~\cite{nam2020lff}; GroupDRO~\cite{sagawa2020groupdro}; FLAC~\cite{sarridis2024flac}; and BPA~\cite{bpa2024}.

\subsection{Results}

\begin{table*}[t]
  \centering
  \scriptsize
  \setlength{\tabcolsep}{3pt}
  \caption{\wga{} (\%) on Set~B for FairFace dataset.}
  \label{tab:mitigation}
  \resizebox{\textwidth}{!}{%
  \begin{tabular}{llccccccc}
    \toprule
    Ratio & Method & FairFace & MiVOLO & ViT-Age & SigLIP2 & MetaCLIP2 & Qwen2.5-VL & LLaVA-1.6 \\
    \midrule
    \multirow{5}{*}{1\%} & ERM & $84.7_{\pm 1.4}$ & $59.2_{\pm 2.7}$ & $80.6_{\pm 4.4}$ & $87.2_{\pm 4.1}$ & $78.0_{\pm 1.3}$ & $85.8_{\pm 0.7}$ & $89.2_{\pm 1.5}$ \\
     & LfF & $84.3_{\pm 1.1}$ & $57.8_{\pm 1.6}$ & $84.0_{\pm 0.6}$ & $88.6_{\pm 3.9}$ & $81.3_{\pm 1.8}$ & $85.2_{\pm 1.0}$ & $90.5_{\pm 2.4}$ \\
     & GroupDRO & $93.7_{\pm 1.1}$ & $71.0_{\pm 1.8}$ & $77.3_{\pm 1.4}$ & $82.7_{\pm 4.0}$ & $86.9_{\pm 1.7}$ & $92.3_{\pm 0.9}$ & $92.8_{\pm 1.2}$ \\
     & FLAC & $87.4_{\pm 2.0}$ & $61.4_{\pm 1.2}$ & $87.2_{\pm 2.3}$ & $87.4_{\pm 2.1}$ & $81.4_{\pm 0.8}$ & $88.5_{\pm 0.9}$ & $92.2_{\pm 2.4}$ \\
     & BPA & $\mathbf{95.6_{\pm 1.5}}$ & $\mathbf{93.0_{\pm 0.6}}$ & $\mathbf{92.3_{\pm 2.7}}$ & $\mathbf{93.2_{\pm 1.7}}$ & $\mathbf{89.2_{\pm 1.2}}$ & $\mathbf{94.0_{\pm 2.0}}$ & $\mathbf{95.4_{\pm 2.0}}$ \\
    \midrule
    \multirow{5}{*}{5\%} & ERM & $87.6_{\pm 0.2}$ & $73.9_{\pm 0.3}$ & $94.0_{\pm 1.3}$ & $95.5_{\pm 0.3}$ & $83.3_{\pm 2.9}$ & $89.2_{\pm 1.4}$ & $96.8_{\pm 0.7}$ \\
     & LfF & $87.3_{\pm 1.9}$ & $66.7_{\pm 4.1}$ & $93.2_{\pm 0.8}$ & $94.7_{\pm 0.8}$ & $84.8_{\pm 0.6}$ & $89.3_{\pm 1.8}$ & $96.0_{\pm 1.3}$ \\
     & GroupDRO & $94.7_{\pm 0.6}$ & $70.4_{\pm 0.8}$ & $77.6_{\pm 2.6}$ & $86.3_{\pm 1.9}$ & $\mathbf{88.2_{\pm 2.6}}$ & $\mathbf{92.6_{\pm 0.8}}$ & $92.2_{\pm 0.6}$ \\
     & FLAC & $89.0_{\pm 1.4}$ & $80.8_{\pm 1.9}$ & $94.1_{\pm 0.5}$ & $96.3_{\pm 0.7}$ & $86.2_{\pm 1.6}$ & $91.8_{\pm 2.0}$ & $\mathbf{97.5_{\pm 1.3}}$ \\
     & BPA & $\mathbf{96.5_{\pm 0.9}}$ & $\mathbf{96.9_{\pm 0.7}}$ & $\mathbf{96.1_{\pm 1.2}}$ & $\mathbf{96.8_{\pm 0.3}}$ & $87.5_{\pm 0.4}$ & $92.5_{\pm 1.4}$ & $93.9_{\pm 4.6}$ \\
    \midrule
    \multirow{5}{*}{10\%} & ERM & $90.7_{\pm 1.0}$ & $81.4_{\pm 1.3}$ & $94.3_{\pm 0.2}$ & $95.4_{\pm 0.8}$ & $88.1_{\pm 1.2}$ & $92.8_{\pm 3.5}$ & $97.9_{\pm 0.7}$ \\
     & LfF & $89.3_{\pm 0.7}$ & $74.7_{\pm 5.7}$ & $94.5_{\pm 0.8}$ & $95.3_{\pm 0.6}$ & $87.6_{\pm 0.8}$ & $93.5_{\pm 1.0}$ & $97.8_{\pm 0.3}$ \\
     & GroupDRO & $\mathbf{93.7_{\pm 2.0}}$ & $70.7_{\pm 1.2}$ & $82.1_{\pm 1.9}$ & $87.3_{\pm 2.0}$ & $86.5_{\pm 2.0}$ & $92.9_{\pm 0.2}$ & $94.5_{\pm 0.9}$ \\
     & FLAC & $91.3_{\pm 0.9}$ & $84.8_{\pm 3.8}$ & $94.8_{\pm 0.9}$ & $95.9_{\pm 0.4}$ & $89.7_{\pm 1.2}$ & $\mathbf{95.8_{\pm 0.9}}$ & $\mathbf{98.7_{\pm 0.2}}$ \\
     & BPA & $92.6_{\pm 3.2}$ & $\mathbf{92.3_{\pm 7.1}}$ & $\mathbf{96.9_{\pm 0.6}}$ & $\mathbf{96.5_{\pm 1.0}}$ & $\mathbf{93.9_{\pm 1.5}}$ & $91.0_{\pm 2.1}$ & $96.5_{\pm 3.3}$ \\
    \bottomrule
  \end{tabular}
  }
\end{table*}
\begin{table}[htbp]
  \centering
  \caption{Flip rate (\%) of BPA at ratio 1\%, with difference vs.\ ERM in parentheses.}
  \label{tab:bpa_vs_erm_001}
  \begin{tabular}{lcccc}
    \toprule
    Model & Beard & Eye Makeup & Lipstick & Mustache \\
    \midrule
    FairFace & $3.1_{\pm 1.7}$ \textcolor{blue}{(-6.2)} & $0.5_{\pm 0.1}$ \textcolor{blue}{(-0.6)} & $1.7_{\pm 0.6}$ \textcolor{blue}{(-1.5)} & $0.5_{\pm 0.3}$ \textcolor{blue}{(-1.9)} \\
    MiVOLO & $3.7_{\pm 0.5}$ \textcolor{blue}{(-30.6)} & $0.9_{\pm 0.5}$ \textcolor{blue}{(-6.2)} & $2.2_{\pm 0.4}$ \textcolor{blue}{(-4.1)} & $0.2_{\pm 0.1}$ \textcolor{blue}{(-0.8)} \\
    ViT-Age & $5.8_{\pm 1.8}$ \textcolor{blue}{(-6.3)} & $0.6_{\pm 0.3}$ \textcolor{blue}{(-0.6)} & $1.3_{\pm 0.6}$ \textcolor{blue}{(-0.7)} & $0.3_{\pm 0.1}$ \textcolor{blue}{(-2.7)} \\
    MetaCLIP2 & $2.8_{\pm 0.7}$ \textcolor{blue}{(-5.5)} & $1.0_{\pm 0.1}$ \textcolor{blue}{(-0.4)} & $4.9_{\pm 0.2}$ \textcolor{blue}{(-3.5)} & $1.4_{\pm 0.3}$ \textcolor{blue}{(-1.6)} \\
    SigLIP2 & $0.7_{\pm 0.4}$ \textcolor{blue}{(-0.8)} & $0.8_{\pm 0.3}$ \textcolor{blue}{(-0.4)} & $3.4_{\pm 0.5}$ \textcolor{blue}{(-2.8)} & $1.0_{\pm 0.8}$ \textcolor{blue}{(-0.4)} \\
    Qwen2.5-VL & $4.0_{\pm 0.9}$ \textcolor{blue}{(-4.5)} & $0.8_{\pm 0.1}$ \textcolor{blue}{(-0.3)} & $4.2_{\pm 0.9}$ \textcolor{blue}{(-2.2)} & $0.5_{\pm 0.3}$ \textcolor{blue}{(-1.4)} \\
    LLaVA-1.6 & $2.2_{\pm 0.5}$ \textcolor{blue}{(-3.1)} & $0.1_{\pm 0.0}$ \textcolor{blue}{(-0.4)} & $0.9_{\pm 0.3}$ \textcolor{blue}{(-2.9)} & $0.3_{\pm 0.0}$ \textcolor{blue}{(-1.1)} \\
    \bottomrule
  \end{tabular}
\end{table}

The results in Table~\ref{tab:mitigation} show that bias mitigation approaches can substantially improve robustness. With only 1\% of manipulated samples, BPA achieves the best \wga{} for all seven backbones, with particularly large gains for the most vulnerable model, MiVOLO, improving from 59.2\% under ERM to 93.0\%. Similar gains are observed for the remaining backbones, indicating that the frozen representations already contain useful information for separating robust age-related characteristics from manipulation-specific attributes.

ERM becomes stronger as the ratio increases, as its robustness scales directly with how many manipulated samples it observes during training. Similarly, increasing the proportion of manipulated training examples generally improves robustness for most mitigation methods, although the gains are not uniform. Overall, BPA remains the most reliable method, yielding the best results across most backbone-ratio combinations. FLAC is also competitive, especially for Qwen2.5-VL and LLaVA-1.6 at 10\%, while GroupDRO performs well for some specific backbones but can degrade substantially for others. 

Focusing on the most challenging setting, namely the 1\% manipulated-sample ratio, Table~\ref{tab:bpa_vs_erm_001} reports the flip rates obtained by the best-performing method, i.e., BPA. 
As shown, BPA substantially reduces flip rates across all models and manipulation types. The largest reductions are observed for the initially most vulnerable cases, such as MiVOLO under beard stubble, where the flip rate decreases by 30.6 percentage points compared to ERM.

Overall, these results suggest that robust age verification under simple appearance manipulations can be improved through lightweight adaptation on top of the pretrained backbones.


\section{Conclusion}
\label{sec:conclusion}

We evaluated seven age verification models across three datasets and four facial manipulations. Despite reasonable performance on the original images,  most models show manipulation-specific vulnerabilities. Drawn beard stubble is the most damaging manipulation overall, producing a 20.8\% average flip rate and reaching 61.4\% for MiVOLO. These vulnerabilities also vary across demographic groups, with female subjects generally more affected than male subjects. 
Mitigation approaches improve robustness, with BPA achieving the best \wga{} across all seven backbones at the 1\% setting.
Note that all manipulations used in this study are simulated rather than physically applied by a real subject. A systematic evaluation on real, physically manipulated photographs remains an important direction for future work. 
Moreover, we evaluate face-based age estimation models in isolation, whereas deployed systems could combine visual predictions with other signals, such as network and account activity. Our results pertain to the robustness of the visual component specifically; whether these manipulation-induced errors survive as bypasses within a full multi-signal system, where other signals could compensate for an incorrect visual estimate, remains an open question. 
Overall, robustness to manipulations and demographic disparities should be evaluated alongside typical performance metrics, while facial age estimation should remain a component of a broader verification process.

\subsubsection{Acknowledgments.} 
This research was supported by the EU Horizon Europe projects ELIAS (grant no. 101120237) and ELLIOT (grant no. 101214398).

%
%
%
\bibliographystyle{splncs04}
\bibliography{main}

\end{document}